\pdfoutput=1

\documentclass[11pt,table]{article}

\usepackage[final]{acl}

\usepackage{times}
\usepackage{latexsym}
\usepackage{pgfplots}
\usepackage{tikz}
\usepackage{amsmath}
\usepackage{listings}

\usepackage[T1]{fontenc}

\usepackage[utf8]{inputenc}

\usepackage{microtype}

\usepackage{inconsolata}
\usepackage{graphicx}
\usepackage{multirow}
\usepackage{booktabs}
\usepackage{graphicx}
\usepackage{amsmath}
\usepackage{subcaption}
\usepackage{paralist}
\newcommand{\ModelName}{UniEDU}

\definecolor{NavyBlue}{rgb}{0.0, 0.0, 0.5}
\definecolor{ForestGreen}{rgb}{0.13, 0.55, 0.13}
\definecolor{Red}{rgb}{1.0, 0.0, 0.0}
\definecolor{Blue}{rgb}{0.0, 0.0, 1.0}
\definecolor{Orange}{rgb}{1.0, 0.647, 0.0}
\definecolor{Purple}{rgb}{0.5, 0.0, 0.5} 
\definecolor{Teal}{rgb}{0.0, 0.5, 0.5}
\definecolor{DarkCyan}{rgb}{0.0, 0.55, 0.55}  
\definecolor{Goldenrod}{rgb}{0.85, 0.65, 0.13}  
\title{\ModelName{}: Toward Unified and Efficient Large Multimodal Models \\ for Educational Tasks}

\author{Zhendong Chu\textsuperscript{\rm $\heartsuit$}\thanks{Equal contribution.},
 Jian Xie\textsuperscript{\rm $\spadesuit$}\footnotemark[1],
 Shen Wang\textsuperscript{\rm $\heartsuit$},
 Zichao Wang\textsuperscript{\rm $\sigma$}\thanks{ZW's contribution to this paper was limited to advising on the task set up and evaluation.  ZW did not participate in the experiments or model development.},
 Qingsong Wen\textsuperscript{\rm $\heartsuit$}\\
\textsuperscript{\rm $\heartsuit$}Squirrel Ai Learning \\
\textsuperscript{\rm $\spadesuit$}Fudan University 
\textsuperscript{\rm $\sigma$}Adobe Research \\
\texttt{\{zc9uy@virginia.edu, qingsongedu@gmail.com\}}
}

\begin{document}
\maketitle
\begin{abstract}
Education materials for K-12 students often consist of multiple modalities, such as text and images, posing challenges for models to fully understand nuanced information in these materials. 
In this paper, we propose a unified and efficient large multimodal model \ModelName{} designed for various educational applications, including knowledge recommendation, knowledge tracing, time cost prediction, and user answer prediction, all within a single model. 
Unlike conventional task-specific models, \ModelName{} offers a unified solution that excels across multiple educational tasks while maintaining strong generalization capabilities. 
Its adaptability makes it well-suited for real-world deployment in diverse learning environments. 
Furthermore, \ModelName{} is optimized for industry-scale deployment by significantly reducing computational overhead---achieving approximately a 3×  increase in efficiency---while maintaining competitive performance with minimal degradation compared to fully fine-tuned models.
This work represents a significant step toward creating versatile AI systems tailored to the evolving demands of education.

\end{abstract}

\section{Introduction}
\label{sec:intro}
The incorporation of artificial intelligence (AI) significantly enhances the quality of K-12 education by enabling more personalized learning experiences, improving student engagement~\citep{chen2024llms,adetayo2024microsoft}, and providing educators with valuable insights to tailor instruction to individual needs~\citep{bhowmik2024evaluation,zheng2025teaching}.
For example, knowledge recommendation systems leverage AI to suggest relevant learning materials based on students' past performance and preferences~\citep{wang2024large,chu2025llm}, while knowledge tracing techniques track students’ understanding over time, allowing for real-time adjustments to learning paths~\citep{li2024extrakt,shen2024revisiting,yang2024content}. 

Despite these advancements, previous research has primarily focused on plain text modality, while real-world K-12 scenarios often involve multimodal data, such as text and images in question stems.
Furthermore, the significant differences between tasks pose a challenge in designing a unified model that can effectively handle diverse input types. 
However, since user profiles remain consistent across tasks, a unified approach could facilitate seamless knowledge transfer among them. 
For example, while knowledge recommendation is typically framed as a ranking problem and knowledge tracing as a binary classification task, both rely on a shared understanding of student learning behaviors and knowledge states. 
These disparities underscore the need for a unified model capable of handling the complexities of multimodal scenarios and supporting diverse task types within the context of educational AI assistance.

Large Multimodal Models (LMMs)~\citep{liu2023llava,liu2023improvedllava,chen2024internvl} emerge as a promising solution due to their proficiency in handling multimodal data. 
Furthermore, by leveraging the flexibility of natural language, LMMs can reframe tasks in a generative format and tailor input descriptions to effectively support a wide range of distinct tasks.
However, the computational cost of processing long input contexts remains a significant challenge. Since user interaction histories often span extended periods---up to 300 interactions in our study, with a maximum length reaching 45,000 tokens---retaining all interactions would substantially increase token costs, thereby escalating both training and inference expenses. 
A detailed analysis of these computational costs is provided in Section~\ref{subsec:eff_analysis}.
While directly truncating such data may risk losing important information, educational user profiles are largely constructed from interaction histories, which makes them more amenable to compression. Unlike other domains that may require precise memory of all events, educational contexts often tolerate approximate representations, as not all historical details are equally critical for capturing learning behaviors and knowledge states \cite{rendle2023reducing, purificato2024user, chu2024improve}.

To address these challenges and accommodate the unique demands of educational settings, we propose \textbf{\ModelName{}}---a unified large multimodal model optimized for efficient deployment in educational assistant systems. 
\ModelName{} compresses student interaction histories into a compact set of tokens for efficient feature extraction and reformulates diverse real-world educational tasks within a generative framework. 
Comprehensive experiments demonstrate that \ModelName{} achieves strong performance across real-world tasks, outperforming task-specific models while delivering approximately a 300\% improvement in computational efficiency.

\section{Related Work}
\label{sec:relatedwork}
To provide students with support tailored to their abilities and preferences, it is essential to develop an effective AI assistant for K-12 learning. At the outset of AI integration in education, improving e-learning quality was a primary focus~\citep{murtaza2022ai, rahayu2022systematic, xu2025ai}, with techniques such as recommendation systems for personalized learning and adaptive learning platforms playing a central role in tailoring educational content to individual student needs~\citep{zaiane2002building, ali2022enabling}. 
While these systems are effective, most are designed for specific tasks, such as knowledge tracing~\citep{li2024extrakt,shen2024revisiting}, and lack generalization across diverse educational contexts.

With the development of LLMs~\citep{openai2022chatgpt,openai2023gpt4,dubey2024llama,yang2024qwen2}, which leverage superior understanding and generation capabilities, e-learning assistants have made significant strides in expanding their generalization. 
These assistants can now serve both as teaching assistants~\citep{xu2024eduagent,guo2024using,abu2024supporting, xu2025multimodal} and student support~\citep{park2024empowering,liu-etal-2024-personality,scarlatos2025exploring, yan2024errorradar}, reducing teachers' workloads while offering personalized responses tailored to each student's needs. 
However, since LLMs are typically trained on general-domain data, they often struggle to adapt to the multimodal inputs and long-context scenarios common in modern e-learning environments. This highlights the need for a unified model capable of effectively handling these complexities. 

\section{Methodology}
\label{sec:method}
\begin{figure}[t]
\centering

    \includegraphics[width=0.95\linewidth]{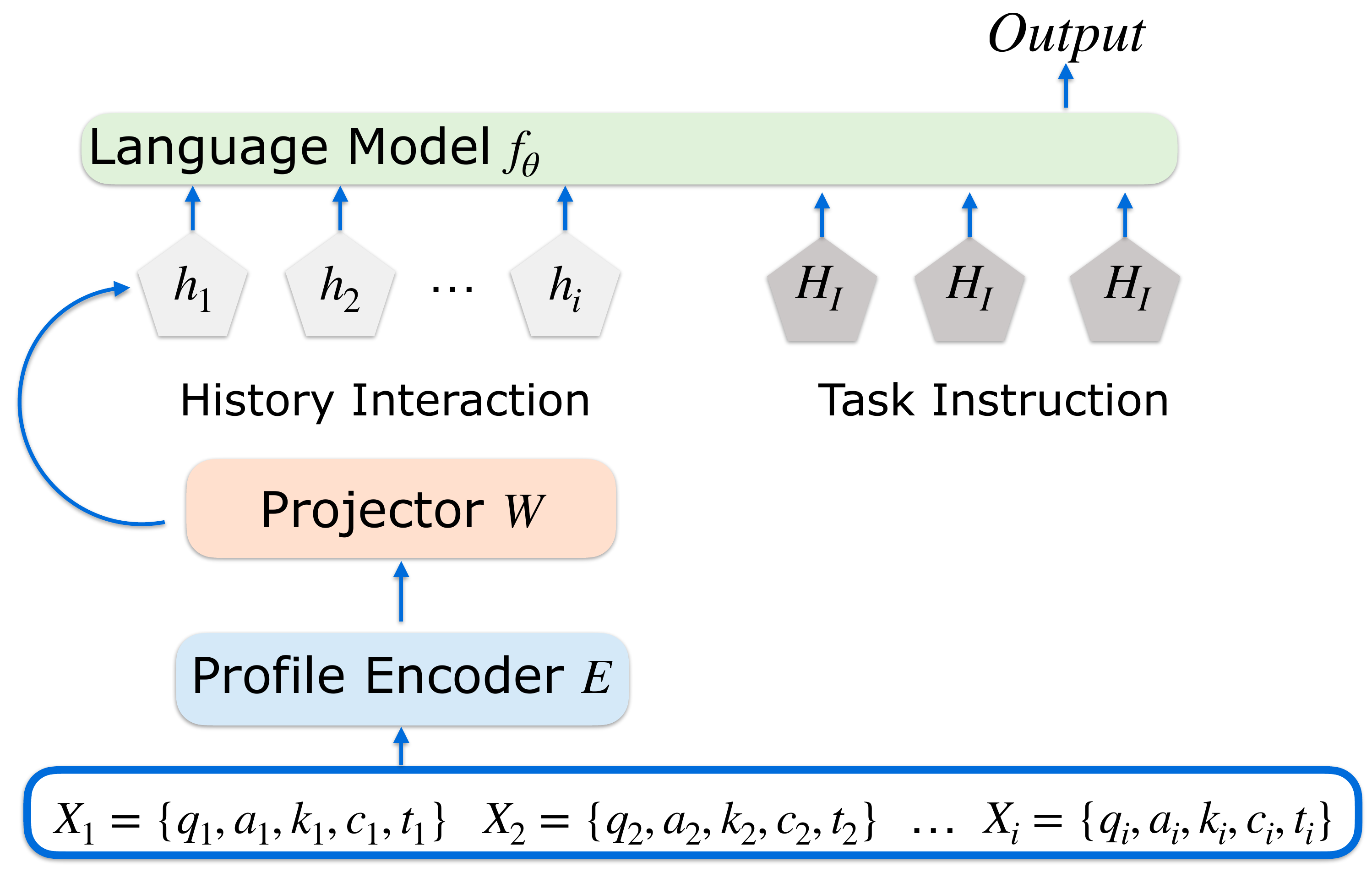}
    \caption{The architecture of \ModelName{}. The profile encoder processes history interactions with multimodal information, while the language model integrates compressed history interactions and task instructions to generate the output.}
    \label{fig:arch}
    \vspace{-1em}
\end{figure}

\subsection{Model Architecture}
\label{subsec:arch}
\ModelName{} comprises two main modules: the \textbf{Profile Encoder} and the \textbf{Language Model}. 
As illustrated in Figure~\ref{fig:arch}, the \textbf{Profile Encoder} is for extracting features from the user's interaction history, while the \textbf{Language Model} is used to generate task-specific responses. 

The \textbf{Profile Encoder} is designed for compressing the user's interaction history into a compact representation that the \textbf{Language Model} can efficiently process.
Formally, let \( S = \{X_{1}, X_{2}, \dots, X_{n}\} \) represent a sequence of user interactions, where each interaction \( X_{i} \) is defined as \( X_{i} = \{q_{i}, a_{i}, k_{i}, c_{i}, t_{i}\} \). 
These components are chosen for their relevance to downstream tasks and are representative of common inputs in real-world educational systems (See Section~\ref{subsec:task}).
Each component of \( X_{i} \) is characterized as follows: \( q_{i} \) denotes the question stem, \( a_{i} \) represents the user's response, \( k_{i} \) corresponds to the knowledge concept associated with the question, \( c_{i} \) indicates the correctness of the user's response, and \( t_{i} \) denotes the time taken by the user to complete the interaction.
\(q_i\) could be multimodal, encompassing both visual information (e.g., figures associated with the question) and textual content.
Please refer to Appendix~\ref{appendix:example} for the demonstration of interaction history.
Given the sequence \( S \), the \textbf{Profile Encoder} compresses it into a feature matrix of shape \( |S| \times m \times h \), where \( |S| \) is the sequence length, \( m \)  is a hyperparameter that determines the number of tokens used to represent each interaction, and \( h \) is the hidden state dimension expected by the language model. 
To ensure compatibility, we apply a \textbf{Projector}---a linear layer that projects the encoder output to the hidden size of the \textbf{Language Model}.

By processing \( S \) through the \textbf{Profile Encoder} and \textbf{Projector}, we obtain \(|S| \times m \) compressed profile embeddings, denoted as \( H_{p} = \{h_{1}, \dots, h_{m|S|}\} \).
These embeddings, together with the uncompressed task instruction embeddings \( H_{i} \), are then processed by the \textbf{Language Model}, which generates responses for different tasks.

\subsection{Training Objective}
To enable multi-task learning, \ModelName{} is trained to generate task-specific outputs conditioned on the user's interaction history and task instructions. In particular, given a sequence of user interactions \( S \) and a task instruction \( \mathbf{X}_{\text{inst}} \), the model generates the corresponding target output \( \mathbf{X}_{\text{t}} \). 
The training process employs the standard auto-regressive training objective, formally defined as:
\begin{equation}
\small
p(\mathbf{X}_{\text{t}} \mid S, \mathbf{X}_{\text{inst}}) = \prod_{i=1}^{|\mathbf{X}_{\text{t}}|} p_{\theta}(x_i \mid S, \mathbf{X}_{\text{inst}}, \mathbf{X}_{\text{t},<i}),
\end{equation}
where \( \theta \) represents the trainable parameters, and \( \mathbf{X}_{\text{inst}} \) and \( \mathbf{X}_{\text{t},<i} \) denote the instruction tokens and the generated target tokens preceding the current prediction token \( x_{i} \), respectively. In Section \ref{sec:experiment}, we discuss the education tasks we considered in detail. 

\begin{table}[t]
    \centering
    \begin{tabular}{c|l}
    \hline
    Variable & Definition \\
    \hline
    \( a \) & number of attention heads \\
    \( b \) & batch size \\
    \( d \) & hidden dimension size \\
    \( l \) & number of transformer layers \\
    \( s \) & sequence length \\
    \( t \) & tensor parallel size \\
    \( v \) & vocabulary size \\
    \hline
\end{tabular}

    \caption{Definitions of the variables.}
    \label{tab:variables}
    \vspace{-1em}
\end{table}

\begin{table*}[t]
    \centering
    \vspace{-2em}
    \begin{tabular}{lcccccc}

\toprule
            & \multirow{2}{*}{Model Size} & \multirow{2}{*}{$VRAM_{para}$} & \multicolumn{3}{c}{$VRAM_{activation}$} & \multirow{2}{*}{$VRAM_{total}$} \\ \cmidrule(l){4-6}
            &                             &                                 & Embedding     & Blocks          & Output        &                                 \\ \midrule
\multicolumn{7}{c}{Training}                                                                                                                            \\ \midrule
Qwen2-VL-2B & 2B                          & 14.9GB                           & 0.3GB      & 61.3GB        & 50.9GB       & 127.4GB                          \\
Qwen2-VL-7B & 7B                          & 52.2GB                          & 0.3GB      & 143.0GB     & 50.9GB    & 246.4GB                       \\
\ModelName-5B        & 5B                          & 37.3GB                          & 1.8MB      & 1.1GB      & 0.4GB      & \textbf{38.8GB}                \\ \midrule
\multicolumn{7}{c}{Inference}                                                                                                                           \\ \midrule
Qwen2-VL-2B & 2B                          & 3.7GB                           & 0.1GB      & 2.2GB        & 25.5GB       & 31.5GB                        \\
Qwen2-VL-7B & 7B                          & 13GB                            & 0.3GB      & 5.1GB     & 25.5GB    & 43.9GB                       \\
\ModelName-5B        & 5B                          & 9.3GB                          & 1.8MB      & 25MB      & 0.2GB      & \textbf{9.5GB}                \\ \bottomrule
\end{tabular}
    \vspace{-0.5em}
    \caption{VRAM Usage Comparison Across Different Models. 
    The results assume \( b = 1 \) and \( t = 1 \), with other parameters set according to their respective models. 
    For \ModelName{}, \( s = 300 \) due to compression, while for the other models, \( s = 45,000 \), representing the maximum number of history interaction tokens. 
    For clarity, token counts for task instructions are omitted, resulting in slight discrepancies in real VRAM usage.
    }
    \vspace{-1.5em}
    \label{tab:vram_comparison}
\end{table*}

\subsection{VRAM Computation}
\label{subsec:vram_comp}
In this section, we provide a detailed analysis of why \ModelName{} is VRAM-efficient for both training and inference. 
We compare \ModelName's VRAM requirements with general fine-tuning demands, focusing on two key components: parameters-loaded VRAM ( $VRAM_{para}$ ) and activation memory ( $VRAM_{activation}$ ).
In Table~\ref{tab:variables}, we list all the definitions of the variables used in this section.

Assuming both model parameters and activations are stored in a 16-bit floating point format, each element requires 2 bytes of storage. 
During the training stage, in addition to \( S_{\text{model}} \) for model loading, additional memory is required for storing optimizer states and gradients.
Specifically, the Adam optimizer maintains two sets of moment estimates---first-order (mean of past gradients) and second-order (variance of past gradients)---for each model parameter, effectively doubling the memory required for optimization. 
As a result, the optimizer states require  $2S_{model}$. 
Additionally, gradient storage requires  $S_{model}$. 
Thus, the total VRAM for loading the parameters in the training stage is:
\begin{equation}
\small
VRAM_{para}^{train} = 4 \times 2 S_{model} = 8 S_{model}.
\end{equation}

During inference, the only VRAM requirement is for loading the model itself, as no optimizer states or gradient storage are needed. 
Therefore, the VRAM required for inference is given by:
\begin{equation}
\small
VRAM_{model}^{\text{infer}} = 2 S_{model}.
\end{equation}

Following the VRAM computation from NVIDIA, the activation memory required for Transformer is given by:
\begin{equation}
\small
VRAM_{activation\text{-}blocks} = \frac{s b d l}{t}\left(34+5 \frac{a s}{d}\right).
\end{equation}

In addition to activations within Transformer blocks, there are activation memory requirements before and after these blocks. The token and position embeddings before the first block require:
\begin{equation}
\small
VRAM_{activation\text{-}embedding} = 4bsd.
\end{equation}

After passing through the Transformer blocks, the output tensors are typically stored in float32, even if the model was loaded at lower precision, as it often casts outputs to float32 by default~\citep{2023vram}. 
During training, probabilities that are the same size as the output tensor also need to be stored, contributing additional memory overhead. 
This results in the following VRAM usage:
\begin{equation}
\small
VRAM_{activation\text{-}output} = 
\begin{cases}
8bsv, & \text{training} \\
4bsv, & \text{inference}
\end{cases}
\end{equation}

To further optimize VRAM consumption, we employ Flash Attention~\citep{dao2022flashattention}, which reduces attention memory complexity from quadratic to linear with respect to the sequence length. 
Given that the sequence length in our setting (up to 45K) is significantly larger than the number of attention heads $a$, the activation memory in the training stage can be approximated as:
\begin{equation}
\small
VRAM_{activation} \approx 34\frac{s b d l}{t} + (4d+8v)bs.
\end{equation}

For inference, the activation memory cost depends on the maximum single activation memory in blocks (i.e., the activation memory of each layer),  as intermediate parameters for updates are not stored. Therefore, the inference cost is:
\begin{equation}
\small
VRAM_{activation} \approx 34\frac{s b d}{t} + 4bsv.
\end{equation}

\subsection{Efficiency Analysis}
\label{subsec:eff_analysis}
In Table~\ref{tab:vram_comparison}, we present the VRAM requirements for Qwen2-VL-2B and 7B~\citep{wang2024qwen2}, as well as our \ModelName{}. 
Due to the long context required for recommendation tasks (up to 45K tokens in our dataset) and the large vocabulary size of modern LLMs, activation memory consumes a substantial amount of VRAM, leading to high computational costs during both training and inference. 
However, after compression, \ModelName{} significantly reduces VRAM usage compared to traditional models, achieving over a 3× reduction even relative to the smaller Qwen2-VL-2B in both training and inference stages. 
This substantial decrease in memory consumption offers a significant advantage for real-world deployment, enabling the model to process larger batches and handle more data within the same timeframe.

\section{Experiments}
\label{sec:experiment}

\subsection{Education Tasks}
\label{subsec:task}
Our system primarily focuses on the subject of mathematics. During training, we use the same user interaction history across tasks while varying task instructions to avoid data leakage. Below, we detail the formulation of each task, with corresponding training examples provided in Appendix~\ref{appendix:example}. 

\noindent\textbf{Knowledge Recommendation.}
This task aims to recommend relevant knowledge concept based on a user's interaction history. 
In the educational assistant context, the model identifies a student's weak areas and provides targeted recommendations, including both foundational knowledge to address weaknesses and advanced knowledge for further development. 
For example, if a student consistently struggles with questions involving quadratic equations, the model may recommend reviewing the fundamentals of factoring and completing the square. 
Conversely, if the student performs well on basic algebraic manipulation, the system might suggest more advanced topics such as functions or inequalities to support continued growth.
Specifically, we define the data format as a triplet \((S, Y, C)\), where \(S\) represents the user's interaction history (as detailed in Section \ref{sec:method}), \(Y\) denotes the ground truth knowledge concepts that reflect the student's weak areas, obtained from real-world exam history.
\(C\) comprises candidate knowledge concepts, including both the ground truth concepts \(Y\) and distractors. 
To construct \(C\), we sample \(K\) candidate concepts, where \(K \in \{5, 10, 25, 50\}\), including one ground truth and \(K - 1\) distractors. The model needs to rank the candidate set 
\(C\) based on its modeling of the student's ability. Task performance is measured using \(\text{Precision@}1\), where a prediction is considered correct only if the ground truth concept is ranked first among the $K$ candidates.

\noindent\textbf{Knowledge Tracing.} This task aims to predict whether a student can correctly answer a given question. The model must capture the user's profile, identifying both strengths and weaknesses, to make an accurate prediction. Specifically, given a sequence of the user's interaction history \(S\) and a question \(Q\), the model is expected to predict a binary outcome: \texttt{[True]} or \texttt{[False]}, where the ground truth is derived from the student's actual answer.
The performance of this task is evaluated based on prediction accuracy. 

\noindent\textbf{Time Cost Prediction.}
The goal of this task is to predict the time a student needs to solve a given question. The task requires the model to understand both the student’s learning path and the inherent difficulty of the task. 
Specifically, similar to Knowledge Tracing, given a sequence of the user's interaction history \(S\) and a question \(Q\), the model is expected to predict an integer value representing the time required, with the ground truth derived from the student's actual time spent. 
We evaluate the model's performance using Mean Absolute Error (MAE).

\noindent\textbf{User Answer Prediction.}
The User Answer Prediction task aims to predict the user's possible answer to a given question based on their interaction history and learning profile. 
If the model thinks that the student can successfully answer the question, it needs to predict the correct answer \cite{liu2022open}. 
However, if the student is unlikely to succeed, the model needs to predict an answer that aligns with the student's profile, reflecting a potentially incorrect response.
This task requires the model to capture the student's strengths, weaknesses, and learning paths to generate realistic answers. 
Specifically, given a sequence of the user's interaction history \(S\) and a question \(Q\), the model predicts the most probable answer, with the ground truth being the student's actual response. 
We use exact match (EM) to evaluate the performance.

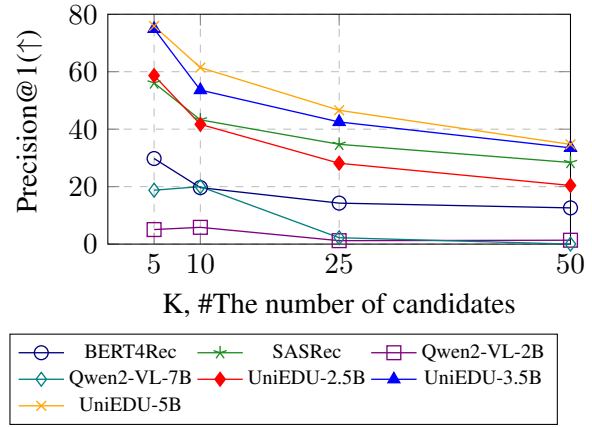
\begin{figure}[t]
    \centering 
    \pgfplotsset{width=\linewidth,height=0.6\linewidth,compat=1.8}
\begin{tikzpicture}
\begin{axis}[
    xmin=0, xmax=50,
    ymin=0, ymax=80,
    xtick={5, 10, 25, 50},
    ytick={0.0, 20.0, 40.0, 60.0, 80.0},
    ymajorgrids=true,
    xmajorgrids=true,
    grid style=dashed,
    xlabel={K, \#The number of candidates},
    ylabel={Precision@1(↑)},
    x label style={at={(axis description cs:0.5,-0.1)},anchor=north,yshift=-6pt},
    y label style={at={(axis description cs:-0.12,0.35)},anchor=south,xshift=16pt},
    legend style={nodes={scale=0.85,font=\footnotesize}, legend columns=3,anchor=north,at={(0.38,-0.32)}, yshift=-6pt}
]
\addplot[
    color=NavyBlue,
    mark=o,
    line width=0.5pt,
    mark size=2.6pt,
    error bars/.cd,
    y dir=both, y explicit,
    error bar style={line width=0.7pt, color=NavyBlue},
    error mark options={rotate=90, NavyBlue, mark size=2.5pt}
    ]
    coordinates {
    (5, 29.78)
    (10, 19.65)
    (25, 14.26)
    (50, 12.60)
    };
    \addlegendentry{BERT4Rec}

\addplot[
    color=ForestGreen,
    mark=star,
    line width=0.5pt,
    mark size=2.6pt,
    error bars/.cd,
    y dir=both, y explicit,
    error bar style={line width=0.7pt, color=ForestGreen},
    error mark options={rotate=90, ForestGreen, mark size=2.5pt}
    ]
    coordinates {
    (5, 56.08)
    (10, 43.24)
    (25, 34.73)
    (50, 28.40)
    };
    \addlegendentry{SASRec}

\addplot[
    color=Purple,
    mark=square,
    line width=0.5pt,
    mark size=2.6pt,
    error bars/.cd,
    y dir=both, y explicit,
    error bar style={line width=0.7pt, color=Purple},
    error mark options={rotate=90, Purple, mark size=2.5pt}
    ]
    coordinates {
    (5, 5.06)
    (10, 5.84)
    (25, 1.19)
    (50, 1.35)
    };
    \addlegendentry{Qwen2-VL-2B}

\addplot[
    color=Teal,
    mark=diamond,
    line width=0.5pt,
    mark size=2.6pt,
    error bars/.cd,
    y dir=both, y explicit,
    error bar style={line width=0.7pt, color=Teal},
    error mark options={rotate=90, Teal, mark size=2.5pt}
    ]
    coordinates {
    (5, 18.75)
    (10, 20.00)
    (25, 2.22)
    (50, 0)
    };
    \addlegendentry{Qwen2-VL-7B}

\addplot[
    color=Red,
    mark=diamond*,
    line width=0.5pt,
    mark size=2.6pt,
    error bars/.cd,
    y dir=both, y explicit,
    error bar style={line width=0.7pt, color=Red},
    error mark options={rotate=90, Red, mark size=2.5pt}
    ]
    coordinates {
    (5, 58.65)
    (10, 41.68)
    (25, 28.13)
    (50, 20.41)
    };
    \addlegendentry{\ModelName-2.5B}

\addplot[
    color=Blue,
    mark=triangle*,
    line width=0.5pt,
    mark size=2.6pt,
    error bars/.cd,
    y dir=both, y explicit,
    error bar style={line width=0.7pt, color=Blue},
    error mark options={rotate=90, Blue, mark size=2.5pt}
    ]
    coordinates {
    (5, 74.90)
    (10, 53.60)
    (25, 42.50)
    (50, 33.50)
    };
    \addlegendentry{\ModelName-3.5B}

\addplot[
    color=Orange,
    mark=x,
    line width=0.5pt,
    mark size=2.6pt,
    error bars/.cd,
    y dir=both, y explicit,
    error bar style={line width=0.7pt, color=Orange},
    error mark options={rotate=90, Orange, mark size=2.5pt}
    ]
    coordinates {
    (5, 75.87)
    (10, 61.43)
    (25, 46.57)
    (50, 34.70)
    };
    \addlegendentry{\ModelName-5B}
\end{axis}
\end{tikzpicture}
    \vspace{-1.3em}
    \caption{Performance comparison of seven models on the Knowledge Recommendation task.} 
    \label{fig:kr_result}
\end{figure}

\subsection{Implementation Details}  
\noindent\textbf{Datasets. }
\label{app:data}
We collect our dataset from real student exercise data on a widely used e-learning platform and construct the training data for each task. The statistics of the dataset are shown in Table \ref{tab:data}. Each student's history sequence exceeding 300 interactions is truncated into multiple segments. As a result, although we processed 13,239 students, the total number of sequences exceeds this count.

\begin{table}[t]
\centering
\begin{tabular}{@{}lc@{}}
\toprule
\# of students   & 13,239 \\ 
\# of knowledge & 8,247 \\ 
\# of questions  & 235,687 \\
\# of interactions       & 3,892,084 \\ \bottomrule
\end{tabular}
\caption{Dataset statistics. }
\label{tab:data}
\vspace{-1em}
\end{table}

\noindent\textbf{Baselines.}
To assess \ModelName{}'s effectiveness, we compare it against two representative baselines for each task. Specifically, for knowledge recommendation, we evaluate two widely adopted models: SASRec~\citep{kang2018self} and Bert4Rec~\citep{sun2019bert4rec}. 
For the knowledge tracing task, we consider extraKT~\citep{li2024extrakt} and reKT~\citep{shen2024revisiting}. 
For time cost prediction, we utilize N-BEATS~\citep{Oreshkin2020N-BEATS} and Prophet~\citep{Prophet}. 
For user answer prediction, a generative task requiring the model to produce responses in natural language, we employ Qwen2-VL-2B and Qwen2-VL-7B \cite{wang2024qwen2}. 
Additionally, these two models, without fine-tuning, are included as baselines for the three aforementioned tasks.

All experiments are conducted using the same training and test sets. 
The maximum length of interaction history is 300, with a maximum token cost of 45,000, based on the Qwen tokenizer.
For baseline models that require indexing specific users and items, we follow their official guidelines to complete this process.

\begin{figure*}[t]
    \centering
    \vspace{-2.5em}
    \begin{subfigure}[t]{.35\textwidth}
        \centering
        
        \pgfplotsset{width=\linewidth,height=0.75\linewidth,compat=1.8}
\begin{tikzpicture}
    \begin{axis}[
        width=5.7cm,
        ybar,
        nodes near coords,
        xtick=\empty,  
        ylabel={Accuracy(↑)},
        bar width=6pt,
        ymin=40, ymax=90,
        axis y line*=left,
        font=\footnotesize,
        axis x line*=bottom,
        enlarge x limits=0.8,
        ymajorgrids=true,
        grid style=dashed,
        legend to name=sharedlegend,
        legend style={
            draw=black,
            fill=white,
            font=\footnotesize,
            at={(0.5,-0.7)},
            anchor=south,
            legend columns=3,
            /tikz/column 2/.style={column sep=1pt}
        },
        legend image code/.code={%
            \draw[fill] (0cm,-0.1cm) rectangle (0.16cm,0.1cm);
        }
    ]
    \addplot[color=NavyBlue, fill=NavyBlue!30] coordinates {(1.2, 84.7)};
    \addlegendentry{extraKT}

    \addplot[color=ForestGreen, fill=ForestGreen!30] coordinates {(1.4, 83.4)};
    \addlegendentry{reKT}

    \addplot[color=Purple, fill=Purple!30] coordinates {(1.6, 48.4)};
    \addlegendentry{Qwen2-VL-2B}

    \addplot[color=Teal, fill=Teal!30] coordinates {(1.8, 55.8)};
    \addlegendentry{Qwen2-VL-7B}

    \addplot[color=Red, fill=Red!30] coordinates {(2.0, 75.6)};
    \addlegendentry{ModelName-2.5B}

    \addplot[color=Blue, fill=Blue!30] coordinates {(2.2, 72.2)};
    \addlegendentry{ModelName-3.5B}

    \addplot[color=Orange, fill=Orange!30] coordinates {(2.4, 74.4)};
    \addlegendentry{ModelName-5B}
    \end{axis}
\end{tikzpicture}
        \caption{Knowledge Tracing}
        \label{fig:kt_result}
    \end{subfigure}
    \begin{subfigure}[t]{.35\textwidth}
        \centering
        \pgfplotsset{width=\linewidth,height=0.75\linewidth,compat=1.8}
\begin{tikzpicture}
    \begin{axis}[
        ybar,
        nodes near coords,
        xtick=\empty,  
        ylabel={MAE(↓)},
        bar width=7pt,
        ymin=10, ymax=50,
        axis y line*=left,
        font=\footnotesize,
        axis x line*=bottom,
        enlarge x limits=0.8,
        ymajorgrids=true,
        grid style=dashed,
        legend to name=sharedlegend,
        legend style={
            draw=black,
            fill=white,
            font=\footnotesize,
            at={(0.5,-0.7)},
            anchor=south,
            legend columns=3,
            /tikz/column 2/.style={column sep=1pt}
        },
        legend image code/.code={%
            \draw[fill] (0cm,-0.1cm) rectangle (0.16cm,0.1cm);
        }
    ]
    \addplot[color=DarkCyan, fill=DarkCyan!80] coordinates {(1.2, 33.8)};
    \addlegendentry{N-BEATS}

    \addplot[color=Goldenrod, fill=Goldenrod!80] coordinates {(1.4, 22.0)};
    \addlegendentry{Prophet}

    \addplot[color=Purple, fill=Purple!30] coordinates {(1.6, 46.7)};
    \addlegendentry{Qwen2-VL-2B}

    \addplot[color=Teal, fill=Teal!30] coordinates {(1.8, 28.6)};
    \addlegendentry{Qwen2-VL-7B}

    \addplot[color=Red, fill=Red!30] coordinates {(2.0, 17.3)};
    \addlegendentry{ModelName-2.5B}

    \addplot[color=Blue, fill=Blue!30] coordinates {(2.2, 20.5)};
    \addlegendentry{ModelName-3.5B}

    \addplot[color=Orange, fill=Orange!30] coordinates {(2.4, 18.6)};
    \addlegendentry{ModelName-5B}
    \end{axis}
\end{tikzpicture}
        \caption{Time Cost Prediction}
        \label{fig:tcp_result}
    \end{subfigure}
    \begin{subfigure}[t]{.25\textwidth}
        \centering
        \vspace{-8em}
        \pgfplotsset{width=\linewidth,height=0.8\linewidth,compat=1.8}
\begin{tikzpicture}
    \begin{axis}[
        width=4cm,
        height=4.2cm,
        ybar,
        nodes near coords,
        xtick=\empty,  
        ylabel={EM(↑)},
        bar width=5pt,
        ymin=1, ymax=15,
        axis y line*=left,
        font=\footnotesize,
        axis x line*=bottom,
        enlarge x limits=0.8,
        ymajorgrids=true,
        grid style=dashed,
        legend to name=sharedlegend,
        legend style={
            draw=black,
            fill=white,
            font=\footnotesize,
            at={(0.5,-0.7)},
            anchor=south,
            legend columns=9,
            /tikz/column 2/.style={column sep=1pt}
        },
        legend image code/.code={%
            \draw[fill] (0cm,-0.1cm) rectangle (0.16cm,0.1cm);
        }
    ]

    \addlegendimage{color=NavyBlue, fill=NavyBlue!30}
    \addlegendentry{extraKT}
    
    \addlegendimage{color=ForestGreen, fill=ForestGreen!30}
    \addlegendentry{reKT}

    \addlegendimage{color=DarkCyan, fill=DarkCyan!80}
    \addlegendentry{N-BEATS}
    
    \addlegendimage{color=Goldenrod, fill=Goldenrod!80}
    \addlegendentry{Prophet}
    
    \addplot[color=Purple, fill=Purple!30] coordinates {(1.2, 1.9)};
    \addlegendentry{Qwen2-VL-2B}

    \addplot[color=Teal, fill=Teal!30] coordinates {(1.4, 5.0)};
    \addlegendentry{Qwen2-VL-7B}

    \addplot[color=Red, fill=Red!30] coordinates {(1.6, 2.7)};
    \addlegendentry{\ModelName-2.5B}

    \addplot[color=Blue, fill=Blue!30] coordinates {(1.8, 10.0)};
    \addlegendentry{\ModelName-3.5B}

    \addplot[color=Orange, fill=Orange!30] coordinates {(2.0, 11.4)};
    \addlegendentry{\ModelName-5B}

\addlegendentry{reKT}
    \end{axis}
\end{tikzpicture}
        \vspace{0.5em}
        \caption{User Answer Prediction}
        \label{fig:ap_result}
    \end{subfigure}
    \vspace{-0.5em}
    \pgfplotslegendfromname{sharedlegend}
    \caption{Performance comparison of \ModelName{} and baseline models on Knowledge Tracing, Time Cost Prediction, and User Answer Prediction.}
    \vspace{-0.5em}
    \label{fig:mix}
\end{figure*}
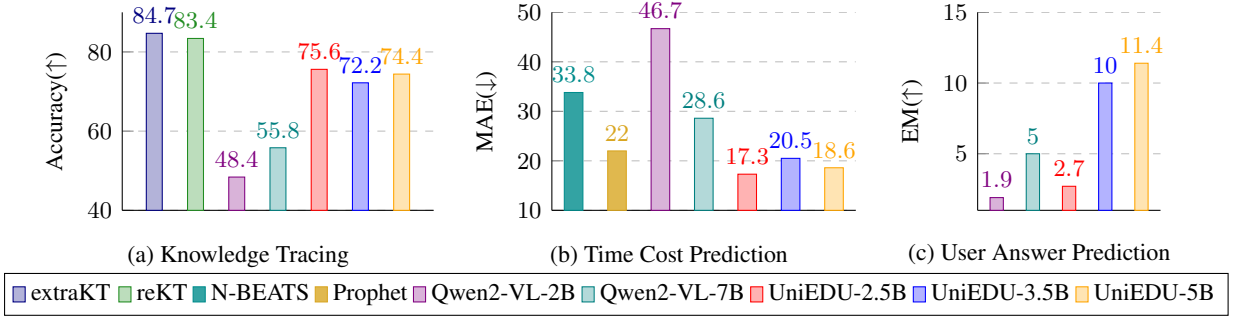

\noindent\textbf{Backbone Models.} We fine-tune \ModelName~ on the four tasks outlined in Section~\ref{subsec:task}. 
For each task, we use 24,504 sequences for training, with 5\% of the data randomly selected as a validation set, and 2,784 sequences for testing. 
For \ModelName, we fix the encoder model as Qwen2-VL-2B and vary the language model size by using Qwen2.5-0.5B, Qwen2.5-1.5B, and Qwen2.5-3B. 
These configurations form \ModelName-2.5B, \ModelName-3.5B, and \ModelName-5B, enabling us to assess the impact of model size on performance.

\subsection{Results}
We compare the performance of baselines and \ModelName{} with various parameter sizes on four tasks.
The results are reported in Figure~\ref{fig:kr_result} and Figure~\ref{fig:mix}. 

First, \textbf{\ModelName{} demonstrates strong performance across four tasks.} Except for knowledge tracing, \ModelName{} outperforms task-specific models in Knowledge Recommendation, Time Cost Prediction, and User Answer Prediction. Notably, it achieves performance gains of approximately 30\% and 20\% over the best baselines in knowledge recommendation and time cost prediction, respectively. While \ModelName{} performs competitively in knowledge tracing, it slightly lags behind specialized models like extraKT and reKT, which are better suited for simpler discriminative tasks. However, these models struggle with unseen items, whereas  \ModelName{} handles them effectively through natural language descriptions.

Second, \textbf{model size significantly affects generative tasks but has limited impact on discriminative ones.} For Knowledge Tracing and Time Cost Prediction, performance remains relatively stable across model sizes. 
In contrast, for Knowledge Recommendation and User Answer Prediction, larger models like \ModelName-5B show clear advantages over smaller variants. This suggests that tasks requiring longer or more complex generation benefit more from increased language model capacity.

\begin{figure}[t]
\centering
    \includegraphics[width=\linewidth]{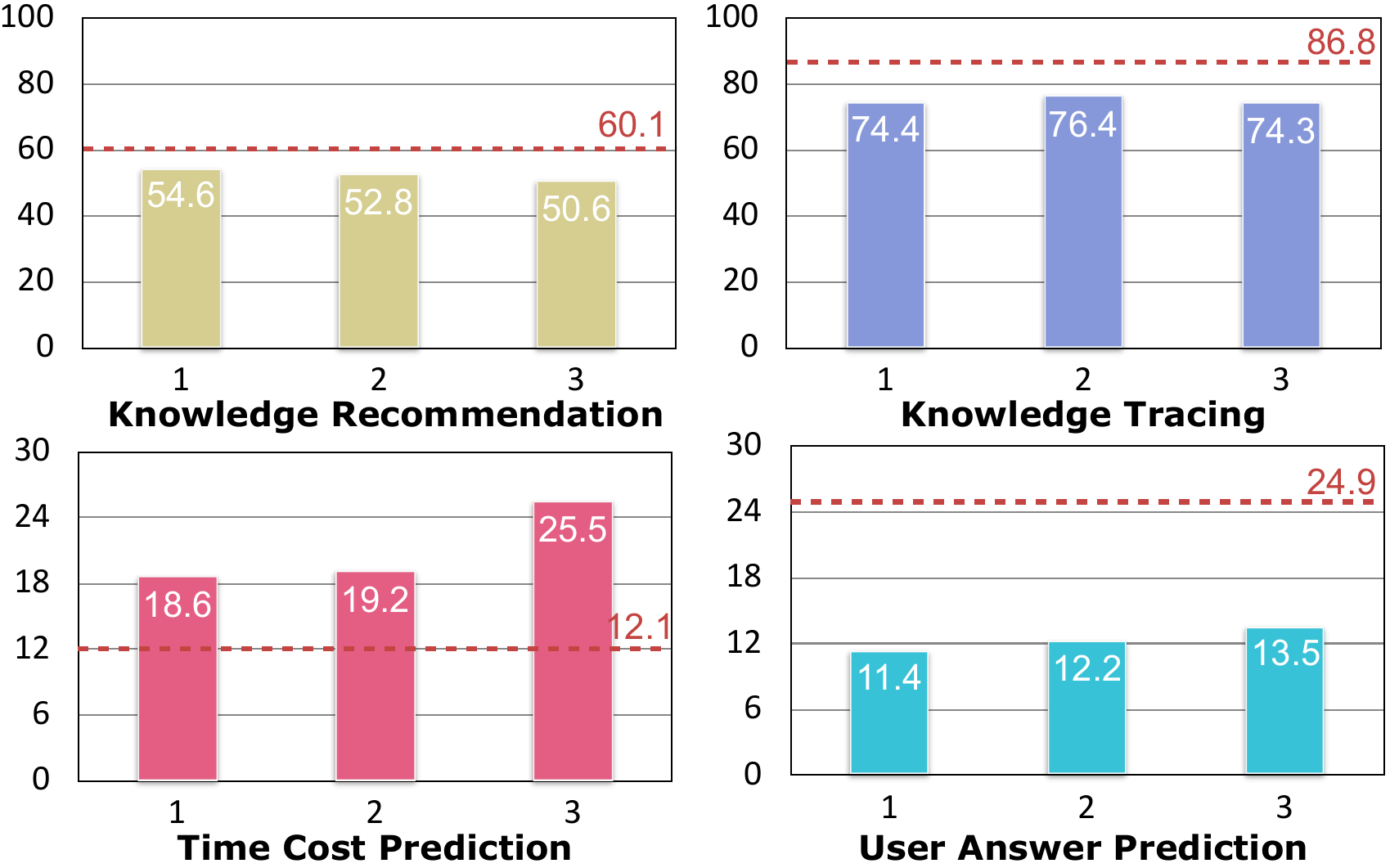}
    \caption{Performance of \ModelName{}-5B with different numbers of compression tokens. The red dashed line indicates Qwen2-VL-2B with full fine-tuning.}
    \label{fig:compratio}
    \vspace{-1.5em}
\end{figure}

\subsection{Analysis of Compression Tokens}
To evaluate the impact of compression ratio on performance, we vary the number of compression tokens (i.e., \( m \), as defined in Section~\ref{subsec:arch}) to 1, 2, 3, meaning each history interaction \( X_{i} \) is compressed into 1, 2, 3 hidden states. 
We conduct this analysis on \ModelName-5B and use Qwen2-VL-2B as an upper bound, excluding the 7B variant due to its high computational cost.

Results in Figure~\ref{fig:compratio} show that increasing the number of compression tokens slightly degrades performance in Knowledge Recommendation and Knowledge Tracing, likely due to noise from excessive historical context. 
In contrast, User Answer Prediction benefits from additional context, as it requires modeling both historical interactions and candidate questions. 
Overall, our compression approach provides substantial efficiency gains with minimal performance loss, except in the more complex generative setting of User Answer Prediction.
Furthermore, compared to the fully fine-tuned model, our compression technique achieves significant improvements in training and inference efficiency with minimal performance degradation, except for the User Answer Prediction task.

\section{Conclusion}
\label{sec:conclusion}
In this paper, we propose \ModelName, a unified generative model for education that effectively handles various multimodal tasks while being computationally efficient. 
Unlike task-specific models, \ModelName{} not only achieves better performance but also generalizes well across different educational challenges, making it suitable for real-world deployment. 
Extensive experiments validate its effectiveness, showing that compared to fully fine-tuned models, \ModelName{} reduces computation costs by approximately $300\%$ while incurring minor performance drops. 
Overall, \ModelName{} represents a promising step toward integrating LMMs into industrial education applications, offering a scalable and efficient approach to personalized learning.

\section*{Limitations}
While \ModelName{} shows strong performance and efficiency across multiple educational tasks, several limitations remain. First, its generalizability beyond mathematics to other subjects and task types (e.g., essay grading) has not been explored. Second, the compression strategy, though effective for reducing VRAM, introduces minor performance drops in complex generative tasks, with trade-offs between efficiency and fidelity requiring further study. Third, the interaction-history-based profile modeling may overlook latent learner traits like motivation or learning style; incorporating richer signals could improve personalization.

\section*{Broader Impact Statement}
\ModelName{} has the potential to significantly improve personalized learning by providing targeted knowledge recommendations based on students’ interaction histories. This can enhance student engagement, support educators in curriculum design, and scale AI-driven education to a wider audience. Furthermore, our computationally efficient design in \ModelName{} makes it accessible to institutions and companies with limited computational resources, while maintaining competitive performance with minimal trade-offs. 

However, training large models on student data poses potential risks to student privacy.  To mitigate these concerns, our dataset is constructed from real student interactions, but all personally identifiable information is strictly anonymized. Only interaction data relevant to learning behaviors is retained, while sensitive details such as names, user IDs, and other personal attributes are carefully masked to ensure privacy and compliance with ethical data usage standards.

\bibliography{acl_latex}

\begin{thebibliography}{41}
\providecommand{\natexlab}[1]{#1}

\bibitem[{Abu-Rasheed et~al.(2024)Abu-Rasheed, Abdulsalam, Weber, and
  Fathi}]{abu2024supporting}
Hasan Abu-Rasheed, Mohamad~Hussam Abdulsalam, Christian Weber, and Madjid
  Fathi. 2024.
\newblock Supporting student decisions on learning recommendations: An
  llm-based chatbot with knowledge graph contextualization for conversational
  explainability and mentoring.
\newblock \emph{arXiv preprint arXiv:2401.08517}.

\bibitem[{Adetayo et~al.(2024)Adetayo, Aborisade, and
  Sanni}]{adetayo2024microsoft}
Adebowale~Jeremy Adetayo, Mariam~Oyinda Aborisade, and Basheer~Abiodun Sanni.
  2024.
\newblock Microsoft copilot and anthropic claude ai in education and library
  service.
\newblock \emph{Library Hi Tech News}.

\bibitem[{Ali et~al.(2022)Ali, Hafeez, Humayun, Jamail, Aqib, and
  Nawaz}]{ali2022enabling}
Sadia Ali, Yaser Hafeez, Mamoona Humayun, Nor Shahida~Mohd Jamail, Muhammad
  Aqib, and Asif Nawaz. 2022.
\newblock Enabling recommendation system architecture in virtualized
  environment for e-learning.
\newblock \emph{Egyptian Informatics Journal}, 23(1):33--45.

\bibitem[{Bhowmik et~al.(2024)Bhowmik, West, Barrett, Zhang, Dai, Sokolikj,
  Southerland, Yuan, and Ke}]{bhowmik2024evaluation}
Saptarshi Bhowmik, Luke West, Alex Barrett, Nuodi Zhang, Chih-Pu Dai, Zlatko
  Sokolikj, Sherry Southerland, Xin Yuan, and Fengfeng Ke. 2024.
\newblock Evaluation of an llm-powered student agent for teacher training.
\newblock In \emph{European Conference on Technology Enhanced Learning}, pages
  68--74. Springer.

\bibitem[{Chen and Leitch(2024)}]{chen2024llms}
Celia Chen and Alex Leitch. 2024.
\newblock Llms as academic reading companions: Extending hci through synthetic
  personae.
\newblock \emph{arXiv preprint arXiv:2403.19506}.

\bibitem[{Chen et~al.(2024)Chen, Wu, Wang, Su, Chen, Xing, Zhong, Zhang, Zhu,
  Lu et~al.}]{chen2024internvl}
Zhe Chen, Jiannan Wu, Wenhai Wang, Weijie Su, Guo Chen, Sen Xing, Muyan Zhong,
  Qinglong Zhang, Xizhou Zhu, Lewei Lu, et~al. 2024.
\newblock Internvl: Scaling up vision foundation models and aligning for
  generic visual-linguistic tasks.
\newblock In \emph{Proceedings of the IEEE/CVF conference on computer vision
  and pattern recognition}, pages 24185--24198.

\bibitem[{Chu et~al.(2025)Chu, Wang, Xie, Zhu, Yan, Ye, Zhong, Hu, Liang, Yu
  et~al.}]{chu2025llm}
Zhendong Chu, Shen Wang, Jian Xie, Tinghui Zhu, Yibo Yan, Jinheng Ye, Aoxiao
  Zhong, Xuming Hu, Jing Liang, Philip~S Yu, et~al. 2025.
\newblock Llm agents for education: Advances and applications.
\newblock \emph{arXiv preprint arXiv:2503.11733}.

\bibitem[{Chu et~al.(2024)Chu, Wang, Zhang, Ji, Wang, and Sun}]{chu2024improve}
Zhendong Chu, Zichao Wang, Ruiyi Zhang, Yangfeng Ji, Hongning Wang, and Tong
  Sun. 2024.
\newblock Improve temporal awareness of llms for domain-general sequential
  recommendation.
\newblock In \emph{ICML 2024 Workshop on In-Context Learning}.

\bibitem[{Dao et~al.(2022)Dao, Fu, Ermon, Rudra, and
  R{\'e}}]{dao2022flashattention}
Tri Dao, Dan Fu, Stefano Ermon, Atri Rudra, and Christopher R{\'e}. 2022.
\newblock Flashattention: Fast and memory-efficient exact attention with
  io-awareness.
\newblock \emph{Advances in Neural Information Processing Systems},
  35:16344--16359.

\bibitem[{Dubey et~al.(2024)Dubey, Jauhri, Pandey, Kadian, Al-Dahle, Letman,
  Mathur, Schelten, Yang, Fan et~al.}]{dubey2024llama}
Abhimanyu Dubey, Abhinav Jauhri, Abhinav Pandey, Abhishek Kadian, Ahmad
  Al-Dahle, Aiesha Letman, Akhil Mathur, Alan Schelten, Amy Yang, Angela Fan,
  et~al. 2024.
\newblock The llama 3 herd of models.
\newblock \emph{arXiv preprint arXiv:2407.21783}.

\bibitem[{Guo et~al.(2024)Guo, Latif, Zhou, Huang, and Zhai}]{guo2024using}
Shuchen Guo, Ehsan Latif, Yifan Zhou, Xuan Huang, and Xiaoming Zhai. 2024.
\newblock Using generative ai and multi-agents to provide automatic feedback.
\newblock \emph{arXiv preprint arXiv:2411.07407}.

\bibitem[{Kang and McAuley(2018)}]{kang2018self}
Wang-Cheng Kang and Julian McAuley. 2018.
\newblock Self-attentive sequential recommendation.
\newblock In \emph{2018 IEEE international conference on data mining (ICDM)},
  pages 197--206. IEEE.

\bibitem[{Korthikanti et~al.(2023)Korthikanti, Casper, Lym, McAfee, Andersch,
  Shoeybi, and Catanzaro}]{korthikanti2023reducing}
Vijay~Anand Korthikanti, Jared Casper, Sangkug Lym, Lawrence McAfee, Michael
  Andersch, Mohammad Shoeybi, and Bryan Catanzaro. 2023.
\newblock Reducing activation recomputation in large transformer models.
\newblock \emph{Proceedings of Machine Learning and Systems}, 5:341--353.

\bibitem[{Li et~al.(2024)Li, Bai, Guo, Zheng, Hou, Zhan, Huang, Liu, Gao, and
  Luo}]{li2024extrakt}
Xueyi Li, Youheng Bai, Teng Guo, Ying Zheng, Mingliang Hou, Bojun Zhan, Yaying
  Huang, Zitao Liu, Boyu Gao, and Weiqi Luo. 2024.
\newblock Extending context window of attention based knowledge tracing models
  via length extrapolation.
\newblock In \emph{ECAI 2024}, pages 1479--1486. IOS Press.

\bibitem[{Liu et~al.(2023{\natexlab{a}})Liu, Li, Li, and
  Lee}]{liu2023improvedllava}
Haotian Liu, Chunyuan Li, Yuheng Li, and Yong~Jae Lee. 2023{\natexlab{a}}.
\newblock Improved baselines with visual instruction tuning.

\bibitem[{Liu et~al.(2023{\natexlab{b}})Liu, Li, Wu, and Lee}]{liu2023llava}
Haotian Liu, Chunyuan Li, Qingyang Wu, and Yong~Jae Lee. 2023{\natexlab{b}}.
\newblock Visual instruction tuning.
\newblock In \emph{NeurIPS}.

\bibitem[{Liu et~al.(2022)Liu, Wang, Baraniuk, and Lan}]{liu2022open}
Naiming Liu, Zichao Wang, Richard Baraniuk, and Andrew Lan. 2022.
\newblock Open-ended knowledge tracing for computer science education.
\newblock In \emph{Proceedings of the 2022 Conference on Empirical Methods in
  Natural Language Processing}.

\bibitem[{Liu et~al.(2024)Liu, Yin, Lin, and Chen}]{liu-etal-2024-personality}
Zhengyuan Liu, Stella~Xin Yin, Geyu Lin, and Nancy~F. Chen. 2024.
\newblock \href {https://doi.org/10.18653/v1/2024.emnlp-main.37}
  {Personality-aware student simulation for conversational intelligent tutoring
  systems}.
\newblock In \emph{Proceedings of the 2024 Conference on Empirical Methods in
  Natural Language Processing}, pages 626--642, Miami, Florida, USA.
  Association for Computational Linguistics.

\bibitem[{Meta(2023)}]{Prophet}
Meta. 2023.
\newblock Prophet.

\bibitem[{Murtaza et~al.(2022)Murtaza, Ahmed, Shamsi, Sherwani, and
  Usman}]{murtaza2022ai}
Mir Murtaza, Yamna Ahmed, Jawwad~Ahmed Shamsi, Fahad Sherwani, and Mariam
  Usman. 2022.
\newblock Ai-based personalized e-learning systems: Issues, challenges, and
  solutions.
\newblock \emph{IEEE access}, 10:81323--81342.

\bibitem[{OpenAI(2022)}]{openai2022chatgpt}
OpenAI. 2022.
\newblock Chatgpt.

\bibitem[{OpenAI(2023)}]{openai2023gpt4}
OpenAI. 2023.
\newblock Gpt-4 technical report.
\newblock \emph{arXiv preprint arXiv:2303.08774}.

\bibitem[{Oreshkin et~al.(2020)Oreshkin, Carpov, Chapados, and
  Bengio}]{Oreshkin2020N-BEATS}
Boris~N. Oreshkin, Dmitri Carpov, Nicolas Chapados, and Yoshua Bengio. 2020.
\newblock \href {https://openreview.net/forum?id=r1ecqn4YwB} {N-beats: Neural
  basis expansion analysis for interpretable time series forecasting}.
\newblock In \emph{International Conference on Learning Representations}.

\bibitem[{Park et~al.(2024)Park, Kim, Lee, Kwon, and Kim}]{park2024empowering}
Minju Park, Sojung Kim, Seunghyun Lee, Soonwoo Kwon, and Kyuseok Kim. 2024.
\newblock Empowering personalized learning through a conversation-based
  tutoring system with student modeling.
\newblock In \emph{Extended Abstracts of the CHI Conference on Human Factors in
  Computing Systems}, pages 1--10.

\bibitem[{Purificato et~al.(2024)Purificato, Boratto, and
  De~Luca}]{purificato2024user}
Erasmo Purificato, Ludovico Boratto, and Ernesto~William De~Luca. 2024.
\newblock User modeling and user profiling: A comprehensive survey.
\newblock \emph{arXiv preprint arXiv:2402.09660}.

\bibitem[{Rahayu et~al.(2022)Rahayu, Ferdiana, and
  Kusumawardani}]{rahayu2022systematic}
Nur~W Rahayu, Ridi Ferdiana, and Sri~S Kusumawardani. 2022.
\newblock A systematic review of ontology use in e-learning recommender system.
\newblock \emph{Computers and Education: Artificial Intelligence}, 3:100047.

\bibitem[{Rendle and Zhang(2023)}]{rendle2023reducing}
Steffen Rendle and Li~Zhang. 2023.
\newblock On reducing user interaction data for personalization.
\newblock \emph{ACM Transactions on Recommender Systems}, 1(3):1--28.

\bibitem[{Scarlatos et~al.(2025)Scarlatos, Baker, and
  Lan}]{scarlatos2025exploring}
Alexander Scarlatos, Ryan~S Baker, and Andrew Lan. 2025.
\newblock Exploring knowledge tracing in tutor-student dialogues using llms.
\newblock In \emph{Proceedings of the 15th International Learning Analytics and
  Knowledge Conference}, pages 249--259.

\bibitem[{Shen et~al.(2024)Shen, Yu, Liu, Liang, Wan, Yang, and
  Sun}]{shen2024revisiting}
Xiaoxuan Shen, Fenghua Yu, Yaqi Liu, Ruxia Liang, Qian Wan, Kai Yang, and
  Jianwen Sun. 2024.
\newblock Revisiting knowledge tracing: A simple and powerful model.
\newblock In \emph{Proceedings of the 32nd ACM International Conference on
  Multimedia}, pages 263--272.

\bibitem[{Smirnov(2023)}]{2023vram}
Alex Smirnov. 2023.
\newblock \href {https://asmirnov.xyz/vram} {Breaking down gpu vram
  consumption}.

\bibitem[{Sun et~al.(2019)Sun, Liu, Wu, Pei, Lin, Ou, and
  Jiang}]{sun2019bert4rec}
Fei Sun, Jun Liu, Jian Wu, Changhua Pei, Xiao Lin, Wenwu Ou, and Peng Jiang.
  2019.
\newblock Bert4rec: Sequential recommendation with bidirectional encoder
  representations from transformer.
\newblock In \emph{Proceedings of the 28th ACM international conference on
  information and knowledge management}, pages 1441--1450.

\bibitem[{Wang et~al.(2024{\natexlab{a}})Wang, Bai, Tan, Wang, Fan, Bai, Chen,
  Liu, Wang, Ge et~al.}]{wang2024qwen2}
Peng Wang, Shuai Bai, Sinan Tan, Shijie Wang, Zhihao Fan, Jinze Bai, Keqin
  Chen, Xuejing Liu, Jialin Wang, Wenbin Ge, et~al. 2024{\natexlab{a}}.
\newblock Qwen2-vl: Enhancing vision-language model's perception of the world
  at any resolution.
\newblock \emph{arXiv preprint arXiv:2409.12191}.

\bibitem[{Wang et~al.(2024{\natexlab{b}})Wang, Xu, Li, Zhang, Liang, Tang, Yu,
  and Wen}]{wang2024large}
Shen Wang, Tianlong Xu, Hang Li, Chaoli Zhang, Joleen Liang, Jiliang Tang,
  Philip~S Yu, and Qingsong Wen. 2024{\natexlab{b}}.
\newblock Large language models for education: A survey and outlook.
\newblock \emph{arXiv preprint arXiv:2403.18105}.

\bibitem[{Xu et~al.(2024)Xu, Zhang, and Qin}]{xu2024eduagent}
Songlin Xu, Xinyu Zhang, and Lianhui Qin. 2024.
\newblock Eduagent: Generative student agents in learning.
\newblock \emph{arXiv preprint arXiv:2404.07963}.

\bibitem[{Xu et~al.(2025{\natexlab{a}})Xu, Zhang, Chu, and
  Wen}]{xu2025multimodal}
Tianlong Xu, Yi-Fan Zhang, Zhendong Chu, and Qingsong Wen. 2025{\natexlab{a}}.
\newblock Multimodal ai teacher: Integrating edge computing and reasoning
  models for enhanced student error analysis.
\newblock \emph{AI Magazine}, 46(3):e70030.

\bibitem[{Xu et~al.(2025{\natexlab{b}})Xu, Zhang, Chu, Wang, and
  Wen}]{xu2025ai}
Tianlong Xu, YiFan Zhang, Zhendong Chu, Shen Wang, and Qingsong Wen.
  2025{\natexlab{b}}.
\newblock Ai-driven virtual teacher for enhanced educational efficiency:
  Leveraging large pretrain models for autonomous error analysis and
  correction.
\newblock In \emph{Proceedings of the AAAI Conference on Artificial
  Intelligence}, volume~39, pages 28801--28809.

\bibitem[{Yan et~al.(2024)Yan, Wang, Huo, Li, Li, Su, Gao, Zhang, Xu, Chu
  et~al.}]{yan2024errorradar}
Yibo Yan, Shen Wang, Jiahao Huo, Hang Li, Boyan Li, Jiamin Su, Xiong Gao,
  Yi-Fan Zhang, Tianlong Xu, Zhendong Chu, et~al. 2024.
\newblock Errorradar: Benchmarking complex mathematical reasoning of multimodal
  large language models via error detection.
\newblock \emph{arXiv preprint arXiv:2410.04509}.

\bibitem[{Yang et~al.(2024{\natexlab{a}})Yang, Yang, Hui, Zheng, Yu, Zhou, Li,
  Li, Liu, Huang et~al.}]{yang2024qwen2}
An~Yang, Baosong Yang, Binyuan Hui, Bo~Zheng, Bowen Yu, Chang Zhou, Chengpeng
  Li, Chengyuan Li, Dayiheng Liu, Fei Huang, et~al. 2024{\natexlab{a}}.
\newblock Qwen2 technical report.
\newblock \emph{arXiv preprint arXiv:2407.10671}.

\bibitem[{Yang et~al.(2024{\natexlab{b}})Yang, Chu, Darwin, Han, Li, Wen,
  Copur-Gencturk, Tang, and Liu}]{yang2024content}
Kaiqi Yang, Yucheng Chu, Taylor Darwin, Ahreum Han, Hang Li, Hongzhi Wen,
  Yasemin Copur-Gencturk, Jiliang Tang, and Hui Liu. 2024{\natexlab{b}}.
\newblock Content knowledge identification with multi-agent large language
  models (llms).
\newblock In \emph{International Conference on Artificial Intelligence in
  Education}, pages 284--292. Springer.

\bibitem[{Zaiane(2002)}]{zaiane2002building}
Osmar~R Zaiane. 2002.
\newblock Building a recommender agent for e-learning systems.
\newblock In \emph{International Conference on Computers in Education, 2002.
  Proceedings.}, pages 55--59. IEEE.

\bibitem[{Zheng et~al.(2025)Zheng, Jiang, Gu, Li, Wang, and
  Zhang}]{zheng2025teaching}
Longwei Zheng, Fei Jiang, Xiaoqing Gu, Yuanyuan Li, Gong Wang, and Haomin
  Zhang. 2025.
\newblock Teaching via llm-enhanced simulations: Authenticity and barriers to
  suspension of disbelief.
\newblock \emph{The Internet and Higher Education}, 65:100990.

\end{thebibliography}

\appendix

\section{Appendix}
\subsection{VRAM Calculation}
In the NVIDIA paper~\citep{korthikanti2023reducing}, the hidden size is increased to $4h$ and then reduced back to $h$ across layers. 
This varies in models like Qwen2-VL-7B, where the hidden size is 1,576, and the intermediate size is 8,960. 
For consistency, we adopt the evaluation strategy provided by NVIDIA, which may introduce a minor discrepancy in the real memory costs for models with different configurations.

\subsection{Example Demonstration}
\label{appendix:example}
We provide examples of the four tasks mentioned in Section~\ref{subsec:task}.
Let \( S = \{X_{1}, X_{2}, \dots, X_{n}\} \) denote a sequence of user interactions, where each interaction is defined as \( X_{i} = \{q_{i}, a_{i}, k_{i}, c_{i}, t_{i}\} \), consisting of the question \( q_{i} \), the user's response \( a_{i} \), associated knowledge \( k_{i} \), response correctness \( c_{i} \), and response time \( t_{i} \). The question \( q_{i} \) may be multimodal, incorporating both textual content and visual elements (e.g., figures).

\noindent Interaction History:
\lstset{
    backgroundcolor=\color[RGB]{245,245,245},
    breaklines=true,
    breakindent=0pt,
    basicstyle=\ttfamily\small,
    frame=trbl,
    frameround = tttt,
}
\begin{lstlisting}
Interaction 1:
Question: The difference between 4.6 and 3.26 is ___ less than their sum.
User's Response: ["6.52"]
Correct: True
Response Time: 61s
Knowledge Concept: Three-step word problems with decimal addition and subtraction


Interaction 2:
Question: Shape A is translated ___ by ___ units to get Shape B.
Image: <image>
User Answer: ["down", "5"]
Correct: True
Response Time: 22s
Knowledge Concept: Identifying the direction and distance of translation

Interaction 3:
Question: Xiao Pang's electricity usage in the first quarter was: 105 kWh, 150 kWh, and 99 kWh. The average monthly electricity usage in the first quarter is ___ kWh.
User Answer: ["118"]
Correct: True
Response Time: 71s
Knowledge Concept: Calculating average basic level

.
.
.

Interaction n:
Question: Teacher Hu rode a bicycle to the library and crossed a 1500-meter bridge. On the way there, it took 300 seconds to cross the bridge, and on the way back, it took 200 seconds. Then the average speed over the round trip on the bridge is ___ meters/second.
User Answer: ["6"]
Correct: True
Response Time: 69s
Knowledge Concept: Calculating average speed round trip
\end{lstlisting}

\noindent Knowledge Recommendation:
\lstset{
    backgroundcolor=\color[RGB]{245,245,245},
    breaklines=true,
    breakindent=0pt,
    basicstyle=\ttfamily\small,
    frame=trbl,
    frameround = tttt,
}
\begin{lstlisting}
Based on the user's past problem-solving history, select 1 knowledge concept from the following list that the user is likely to make mistakes on.

Candidate Knowledge Concepts:
Properties of opposite numbers, absolute values, and reciprocals-evaluating algebraic expressions, Solving average problems using equations, Decomposition and composition of numbers within 10 (including 0), Midpoint of a line segment-identifying relationships involving sum, difference, and multiples, Applications of ratio-given the total, Figures obtained by rotation followed by translation, Translation of figures in coordinate systems, Calculations of the form (\alpha + \beta)/2, Applications of linear equations-relationships between points on a number line, Calculating averages-pie chart interpretation, Finding the value represented by a point given the distance between two points, Simplified decimal addition and subtraction, Weighted averages-weights in percentage form, Identifying rotated figures-pattern problems, Weighted averages-weights in ratio form, Corresponding elements of congruent triangles-methods of geometric transformation, Translation drawing on a grid, Translating the rectangular coordinate system, Applications of common factors-finding the number in each group, Word problems with two-digit divisors-constant total amount
\end{lstlisting}

\noindent Knowledge Tracing:
\lstset{
    backgroundcolor=\color[RGB]{245,245,245},
    breaklines=true,
    breakindent=0pt,
    basicstyle=\ttfamily\small,
    frame=trbl,
    frameround = tttt,
}
\begin{lstlisting}
Based on the user's past problem-solving history, determine whether the following question can be answered correctly.
Question: Based on the picture, write two complete equations: ___ (separate with a comma)
Knowledge Concept: Addition and subtraction within 10
Image: <image>
\end{lstlisting}

\noindent Time Cost Prediction
\lstset{
    backgroundcolor=\color[RGB]{245,245,245},
    breaklines=true,
    breakindent=0pt,
    basicstyle=\ttfamily\small,
    frame=trbl,
    frameround = tttt,
}
\begin{lstlisting}
Based on the user's past problem-solving history, estimate how long the user will take to answer the following question (in seconds).
Question: The area of the figure below is ___ square meters.
Knowledge Concept: Area units comparing sizes
Image: <image>
\end{lstlisting}

\noindent User Answer Prediction:
\lstset{
    backgroundcolor=\color[RGB]{245,245,245},
    breaklines=true,
    breakindent=0pt,
    basicstyle=\ttfamily\small,
    frame=trbl,
    frameround = tttt,
}
\begin{lstlisting}
Based on the user's past problem-solving history, predict the answer the user is likely to give for the following question.
Question: Look at the picture and write the equation. The complete equation is ___.
Knowledge Concept: Advanced addition with carrying-adding to 7
Image:<image>
\end{lstlisting}

\subsection{Training Details}
In Table \ref{hyperparameter}, we introduce the hyperparameter configuration used to train \ModelName{-5B}.

\begin{table}[t]
\centering
\begin{tabular}{@{}ll@{}}
\toprule
Hyperparameter                                    & Value                                          \\ \midrule
\rowcolor[HTML]{FFFDFA} 
{\color[HTML]{333333} Encoder Layers}             & {\color[HTML]{333333} 28}                      \\
\rowcolor[HTML]{FFFDFA} 
{\color[HTML]{333333} Encoder Heads}              & {\color[HTML]{333333} 12}                      \\
\rowcolor[HTML]{FFFDFA} 
{\color[HTML]{333333} Encoder Hidden Size}        & {\color[HTML]{333333} 1536}                    \\
\rowcolor[HTML]{FFFDFA} 
{\color[HTML]{333333} Projector Hidden Size}      & {\color[HTML]{333333} 1536-\textgreater{}2048} \\
\rowcolor[HTML]{FFFDFA} 
{\color[HTML]{333333} Language Model Layers}      & {\color[HTML]{333333} 36}                      \\
\rowcolor[HTML]{FFFDFA} 
{\color[HTML]{333333} Language Model Heads}       & {\color[HTML]{333333} 16}                      \\
\rowcolor[HTML]{FFFDFA} 
{\color[HTML]{333333} Language Model Hidden Size} & {\color[HTML]{333333} 2048}                    \\
\rowcolor[HTML]{FFFDFA} 
{\color[HTML]{333333} Max History Window}         & {\color[HTML]{333333} 300}                     \\
\rowcolor[HTML]{FFFDFA} 
{\color[HTML]{333333} \# Compression Tokens}                          & {\color[HTML]{333333} 1/2/3}                   \\
\rowcolor[HTML]{FFFDFA} 
{\color[HTML]{333333} Optimizer}                  & {\color[HTML]{333333} AdamW}                   \\
\rowcolor[HTML]{FFFDFA} 
{\color[HTML]{333333} Learning Rate}              & {\color[HTML]{333333} 2.0e-6}                  \\
\rowcolor[HTML]{FFFDFA} 
{\color[HTML]{333333} Scheduler}                  & {\color[HTML]{333333} Cosine}                  \\
\rowcolor[HTML]{FFFDFA} 
{\color[HTML]{333333} Batch Size per GPU}         & {\color[HTML]{333333} 4}                       \\
\rowcolor[HTML]{FFFDFA} 
{\color[HTML]{333333} Training Steps}             & {\color[HTML]{333333} 6000}                    \\ \bottomrule
\end{tabular}
\caption{Hyperparameter setting for \ModelName{-5B}.}
\label{hyperparameter}
\end{table}

\end{document}